\documentclass[11pt, a4paper]{article}
\usepackage{latexsym}
\usepackage[fleqn]{amsmath}
\usepackage{graphicx}
\usepackage[fleqn]{amsmath}
\usepackage[varg]{txfonts}
\usepackage[english]{babel}
\usepackage[hyperindex=false,
    pageanchor=false,
    pdfmenubar=false,
    pdfpagelabels=false, breaklinks=true]{hyperref}
\usepackage{apacite} 
\usepackage{multirow}
\usepackage{setspace}
\usepackage{epstopdf}
\usepackage{setspace}
\usepackage{threeparttable}

\newcommand{\vc}{\mathbf} 

\newcommand{\mc}{\mathcal}
\newcommand{\mr}{\mathrm}

\newtheorem{thm}{Theorem}[section]

\begin{document}

\title{\textbf{Gene Shaving  using influence function  of  a  kernel method}} 
\author{\textbf{ Md. Ashad Alam$^1$, Mohammad Shahjaman$^2$ and  Md. Ferdush Rahman$^3$}  \\
$^1$Department of Statistics\\ Hajee Mohammad Danesh Science and Technology  University\\ Dinajpur 5200, Bangladesh\\
$^2$Department of Statistics,
$^3$Department of Marketing\\
Begum Rokeya University\\
Rangpur 5400, Bangladesh
}

\date{}
\maketitle
\begin{abstract}
Identifying significant subsets of the genes, gene shaving  is an essential and challenging issue for biomedical research for a huge number of genes and the complex nature of biological networks,. Since positive definite kernel based methods on genomic information  can improve the prediction of diseases, in this paper we proposed a new method, "kernel  gene shaving (kernel canonical correlation  analysis (kernel CCA) based gene shaving). This problem is addressed using the influence function of the kernel CCA.  To investigate  the performance of the proposed method in a comparison of three popular gene selection methods (T-test, SAM and LIMMA), we were  used extensive simulated  and real microarray gene expression datasets. The
performance measures AUC was computed for each of the
methods.  The achievement of the proposed method has improved
than the three well-known gene selection methods. In real data analysis, the proposed method identified a subsets of  $210$ genes out of $2000$ genes.  The network of these genes has significantly more  interactions than expected, which indicates that they may function in a concerted effort  on  colon cancer. 
\end{abstract}  
keywords: Gene shaving, Sensitivity analysis,  Positive-definite kernel, Statistical  machine learning.
 
\section{Introduction}
Gene shaving (GS), identifies subsets of genes,   is an  important research area in the analysis of  an DNA microarray gene expression  data  for biomedical discovery. It leads to gene discovery relevant for a particular target annotation. GS  is not relevant to the  hierarchical clustering and other widely used methods for analyzing gene expression in the genome-wide association studies. GS leads to gene discovery relevant for a specific target annotation. Hence, those selected genes play an important role in the analysis of gene expression data since they are able to differentiate samples from different populations. Despite their successes, these studies are often hampered by their relatively low reproducibility and nonlinearity\cite{Hastei-00, Ruan-11, XChen-12,Castellanos-15}.

The incorporation of various statistical machine learning methods into genomic analysis is a rather recent topic. Since large-scale DNA microarray data present significant challenges for statistical  data analysis as the high dimensionality of genomic features makes the classical approaches framework no longer feasible.  The   kernel methods is a appropriate tools to deal such datasets that map data from a high dimensional space  to a feature space using a nonlinear feature map. The main  advantage of these methods is  to combine statistics and geometry in an effective way \cite{Hofmann-08, Ashad-14, Charpiat-15}. Kernel canonical correlation analysis (kernel CCA) have been extensively studied for decades \cite{Akaho,Ashad-15,Ashad-13},.  

   Nowadays, sensitivity, influence function (IF), based methods have been used to detect an influence observation.   a visualization method for detecting influential observations using the IF of kernel PCA has been propposed   Debruyne et al. (2009) \cite{Debruyne-10}. Filzmoser et al. (2008) also  developed a method for outlier identification in high dimensions \cite{Filzmoser-08}. However, these methods are limited to a single data set. Due to the properties of eigen-decomposition, kernel CCA  and its variant are still a well used method for an biomedical data analysis\cite{Ashad-08,Alam-2016ACMa, Alam-18b}. 

\par The contribution of this paper is three-fold. First, we address the IF of kernel CCA. Second, we use the distribution based methods to confirm the influential observations. Finally, the proposed method is applied to identify a set of gene in both synthesized and real an DNA microarray gene expression  data.

\par The remainder of the paper is organized as follows. In the next section, we provide a brief review of positive definite kernel,  kernel CCA and    IF of kernel CCA.  The utility of the proposed method is demonstrated by  both simulated and real data analysis from an imaging genetics study in Section \ref{Sec:Exp}. In Section \ref{Sec:con}, we summarize our findings and give a perspective for future research.

\section{Method}
\label{sec:methods}
\subsection{Positive definite kernel}
 In kernel methods, a nonlinear feature map is defined by  positive definite kernel. It is known \cite{Aron-RKHS} that a positive definite kernel $k$ is associated with a Hilbert space $\mc{H}$, called  reproducing kernel Hilbert space (RKHS), consisting of functions on $\mathcal{X}$ so that the function value is reproduced by the kernel. For any function $f\in \mc{H}$ and  a point $X\in\mathcal{X}$, the function value $f(X)$ is $
f(X)=\langle f(\cdot),k(\cdot, X)\rangle_{\mc{H}},$ where $\langle ,\rangle_{\mc{H}}$ in the inner product of $\mc{H}$  is called the reproducing property.  Replacing $f$ with $k(\cdot,\tilde{X})$ yields $
k(X,\tilde{X})=\langle k(\cdot, X), k(\cdot, \tilde{X})\rangle_{\mc{H}}$ for any $X,\tilde{X} \in \mc{X}$. A symmetric kernel $k(\cdot, \cdot)$ defined on a space $\mathcal{X}$ is called  positive definite, if  for an  arbitrary number of points $X_1,X_2\ldots, X_n\in\mathcal{X}$ the Gram matrix $(k(X_i, Y_j))_{ij}$ is positive semi-definite.  To transform data for extracting nonlinear features, the mapping $\vc{\Phi}: \mc{X}\to \mc{H}$ is defined as $\vc{\Phi}(X)=k(\cdot, X),$
which is  a function of the first argument. This map is called the  f feature map, and the vector $\vc{\Phi}(X)$ in $\mc{H}$ is called  the  feature vector.  The inner product of two feature vectors is then $\langle \vc{\Phi}(X) ,\vc{\Phi}(\tilde{X} )\rangle_{\mc{H}}=k(X, \tilde{X}).$
This is known as the kernel trick. By this trick the kernel can evaluate the inner product of any two feature vectors efficiently without knowing an explicit form of $\vc{\Phi}(\cdot)$ \cite{Hofmann-08, Ashad-14, Charpiat-15}.

\subsection{Kernel canonical correlation analysis}
\label{sec:CKCCA}
Kernel CCA has been proposed as a nonlinear extension of linear CCA \cite{Akaho}. Researchers have extended the standard kernel CCA  with  an efficient computational algorithm \cite{Back-02}.  Over the last decade,   kernel CCA has been used for various tasks \cite{Alzate2008,Huang-2009, Richfield-17, Ashad-15}. 
Given two sets of random variables $X$  and $Y$ with  two  functions in the RKHS, $f_{X}(\cdot)\in \mc{H}_X$  and  $f_{Y}(\cdot)\in \mc{H}_Y$, the optimization problem of  the random variables $f_X(X)$ and $f_Y(Y)$ is
\begin{eqnarray}
\label{ckcca1}
\rho =\max_{\substack{f_{X}\in \mc{H}_X,f_{Y}\in \mc{H}_Y \\ f_{X}\ne 0,\,f_{Y}\ne 0}}\mr{Corr}(f_X(X),f_Y(Y)).
\end{eqnarray}
The optimizing functions $f_{X}(\cdot)$ and $f_{Y}(\cdotp)$ are determined up to scale.

Using a finite sample, we are able to estimate the desired functions. Given an i.i.d sample, $(X_i,Y_i)_{i=1}^n$ from a joint distribution $F_{XY}$, by taking the inner product with elements or ``parameters" in the RKHS, we have features
$f_X(\cdot)=\langle f_X, \Phi_X(X)\rangle_{\mc{H}_X}= \sum_{i=1}^na_X^ik_X(\cdot,X_i) $ and
 $f_Y(\cdot)=\langle f_Y, \phi_Y(Y)\rangle_{\mc{H}_Y}=\sum_{i=1}^na_Y^ik_Y(\cdot,Y_i)$, where $k_X(\cdot, X)$ and $k_Y(\cdot, Y)$ are the associated kernel functions for $\mc{H}_X$ and $\mc{H}_Y$, respectively. The kernel Gram matrices are defined as   $\vc{K}_X:=(k_X(X_i,X_j))_{i,j=1}^n $ and $\vc{K}_Y:=(k_Y(Y_i,Y_j))_{i,j=1}^n $.  We need the centered kernel Gram matrices $\vc{M}_X=\vc{C}\vc{K}_X\vc{C}$ and $\vc{M}_Y=\vc{C}\vc{K}_Y\vc{C}$, where $\vc{C} = \vc{I}_n -\frac{1}{n}\vc{B}_n$ with $\vc{B}_n = \vc{1}_n\vc{1}^T_n$ and $\vc{1}_n$ is the vector with $n$ ones. The empirical estimate of Eq. (\ref{ckcca1}) is then given by
\begin{eqnarray}
\label{ckcca6}
\hat{\rho}=\max_{\substack{f_{X}\in \mc{H}_X,f_{Y}\in \mc{H}_Y \\ f_{X}\ne 0,\,f_{Y}\ne 0}}\frac{\widehat{\rm{Cov}}(f_X(X),f_Y(Y))}{[\widehat{\rm{Var}}(f_X(X))]^{1/2}[\widehat{\rm{Var}}(f_Y(Y))]^{1/2}}, \nonumber
\end{eqnarray}
where
\begin{align*}
& \widehat{\rm{Cov}}(f_X(X),f_Y(Y))
= \frac{1}{n} \vc{a}_X^T\vc{M}_X\vc{M}_Y \vc{a}_Y \\
& \widehat{\rm{Var}}( f_X(X))
=\frac{1}{n} \vc{a}_X^T\vc{M}_X^2 \vc{a}_X \,  \\ &\widehat{\rm{Var}}( f_Y(Y))=\frac{1}{n} \vc{a}_Y^T\vc{M}_Y^2 \vc{a}_Y,
\end{align*}
where $\vc{a}_{X}$ and $\vc{a}_{Y}$ are the directions of $X$ and $Y$, respectively.

\subsection{Influence function of the  kernel canonical correlation analysis}
\label{sec:IFKCCA}
By using  the   IF   of  kernel PCA, linear PCA and   linear CCA,   we  can derive  the IF of kernel CCA (kernel CC and kernel CVs).
For simplicity, let us define $ \tilde{f}_X(X)=\langle f_X,  \tilde{k}_X (\cdot, X)$.

\begin{thm}
\label{TIFKCCA}
 Given two sets of random variables $(X, Y)$ having the  distribution  $F_{XY}$ and the  { \it j}-th kernel CC ( $\rho_j$) and kernel  CVs ($f_{jX}(X)$ and $f_{jX}(Y)$), the  influence functions of  kernel CC  and kernel CVs  at $Z^\prime = (X^\prime, Y^\prime)$ are
\begin{multline}
\rm{IF} (Z^\prime, \rho_j^2)= - \rho_j^2 \tilde{f}_{jX}^2(X^\prime) + 2 \rho_j \tilde{f}_{jX}(X^\prime) \tilde{f}_{jY}(Y^\prime)  - \rho_j^2 \tilde{f}_{jY}^2(Y^\prime), \nonumber
\end{multline}
\end{thm}
 The above theorem  has been  proved on the basis of previously established ones, such as the IF of  linear  PCA \cite{Tanaka-88, Tanaka-89}, the IF of linear CCA \cite{Romanazii-92},  and the IF of  kernel PCA, respectively. The details proof is given in \cite {Alam-18b}.

 Using the above  result,  we can establish some properties of kernel CCA: robustness, asymptotic consistency and its standard error. In addition, we  are able to identify a set of  genes based on the influence of the data.

For a sample data, let $(X_i, Y_i)_{i=1}^n$ be a sample from the  empirical joint distribution $F_{nXY}$. The EIF (IF based on empirical distribution) of kernel CC and  kernel CVs  at $(X^\prime, Y^\prime)$ for all points $ (X_i, Y_i)$ are $\rm{EIF} (X_i, Y_i, X^\prime, Y^\prime, \rho_j^2) = \widehat{\rm{IF}} ( X^\prime, Y^\prime,  \hat{\rho}_j^2)$, $\rm{EIF} (X_i, Y _i,  X^\prime, Y^\prime, f_{jX}) = \widehat{\rm{IF}} (\cdot, X^\prime, Y^\prime, f_{jX})$, and $\rm{EIF} (X_i, Y_i,   X^\prime, Y^\prime, f_{jY}) = \widehat{\rm{IF}} (\cdot,  X^\prime, Y^\prime, \widehat{f}_{jY})$, respectively.

For the bounded kernels, the IFs defined in Theorem \ref{TIFKCCA} have  three properties: gross error sensitivity, local shift sensitivity, and rejection point. But for unbounded kernels, say a linear, polynomial and so on,  the IFs are not bounded.

\section{Experiments}
\label{Sec:Exp}
To demonstrate the performance of the proposed method in a comparison of three popular gene selection methods (T-test, SAM and LIMMA), we used both simulated and real microarray gene expression datasets. We used three R packages of other methods such as stats, samr and limma. The performance measures AUC were computed for each of the methods using ROC package. All R packages are available in the comprehensive R archive network (cran) or bioconductor.

\subsection{Simulation study}
To investigate the performance of the proposed method in comparison with three popular methods as mentioned above with k= 2 groups, we considered gene expression profiles from both normal distribution and t-distribution. We also considered datasets for both small-and-large-sample cases with different percentages of DE genes.
\subsection{Simulated gene expression profiles generated from Normal Distribution}
The following one-way ANOVA model was used to generate simulated datasets from normal distribution
\begin{eqnarray}
\label{eqsi1}
x_{ijk} = \mu_{ik}+ \epsilon_{ijk}; \nonumber\\ (i=1,2, \cdots, G; j= 1,2, \cdots, n_k; k=1, 2,  \cdots, m)
\end{eqnarray}

where $x_{ijk}$, i is the expression of the $i$th gene for the $j$th samples in k group, $\mu_{ik}$ is the mean of all expressions of ith gene in the kth group and $\epsilon_{ijk}$ is the random error which  usually follows a normal distribution with mean  zero mean and variance $\sigma^2$.

To investigate the performance of the proposed method in a comparison of other three popular methods as early mentioned for $k=2$ groups, we generated $100$ datasets using $100$ times of simulations for both small $(n_1=n_2=3)$ and large $(n_1=n_2=15)$ sample cases using  Eq. (\ref{eqsi1}). The means and the common variance of both groups were set as $(\mu_{i1}, \mu_{i2})\in (3, 5)$ and $\sigma^2=0.1$, respectively. Each dataset for each case represented the gene expression profiles of $G= 1000$ genes, with $n = (n1+n2)$ samples. The proportions of DE gene (pDEG) were set to $0.02$ and $0.06$ for each of the $100$ datasets. We computed average values of different performance measures such as TPR, TNR, FPR, FNR, MER, FDR and AUC based on $20$ and $60$ estimated DE genes by four methods (T-test, SAM, LIMMA and Proposed) for each of $100$ datasets. Fig. 1a and Fig.1b represents the ROC curve based on $20$ estimated DE genes by four methods for both small-and-large-sample cases, respectively. From this figure we observe that the proposed method performed better than other three methods for small-sample case (see Fig.1a). On the other hand, for large-sample case (see Fig.1b) proposed method keeps almost equal performance with other three methods (T-test, SAM and LIMMA).  Fig.2 shows the boxplot of AUC values based on 100 simulated datasets estimated by each of the four methods both for small-and-large-sample cases, respectively. Fig.2a and Fig.2b represent the boxplots of AUC values with pDEG = $0.02$ and $0.06$, respectively. From these boxplots we obtained similar results like ROC curve for every pDEG values. We also noticed that the performance of the methods increases when we increase the value of pDEG $0.02$ to $0.06$. Furthermore, we estimated the average values of different performance measures such TPR, TNR, FPR, FNR, MER, FDR and AUC based on $20$ (pDEG=$0.02$) and $60$ (pDEG=$0.06$) estimated DE genes by each of the methods. The results are summarized in Table 1. In this table the results without and within the brackets (.) indicates average of different performance measures estimated by different methods for small-and-large sample cases, respectively. From this Table 1 we also revealed similar interpretations like ROC curve and boxplots.

 \begin{figure}
\begin{center}
\includegraphics[width=8cm, height=8cm]{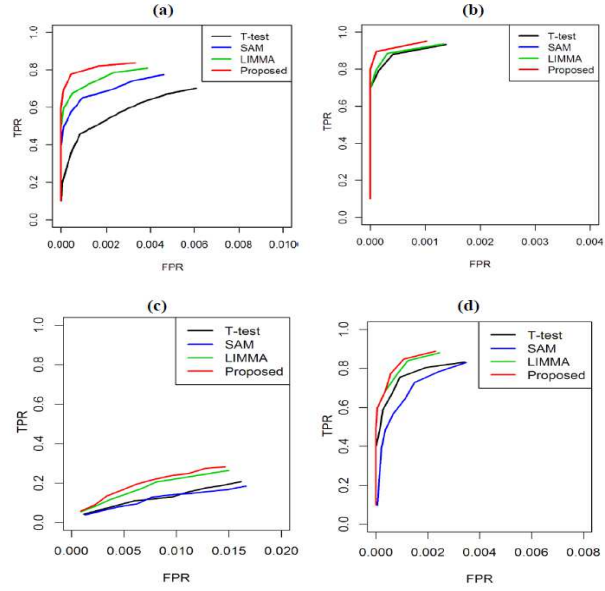}
\caption{ Performance evaluation using ROC-curve produced by the four methods (T-test, SAM, LIMMA and Proposed) based on 100 datasets with pDEG=0.02. Datasets were generated from normal distribution for (a) and (b) and datasets were generated from t-distribution for (c) and (d), where (a) and (c) represents ROC curve for small-sample case (n1=n2=3) and (b) and (d) represents ROC curve for large-sample case (n1=n2=15).}
\label{fig:roc1}
\end{center}
\end{figure}

\begin{table}[]
\caption{Performance evaluation of different methods based on simulated gene expression dataset generated from normal distribution.}
\begin{tabular}{|c|c|c|c|c|c|c|c|}
\hline
\multirow{2}{*}{\textbf{Methods}} & \multicolumn{7}{c|}{\textbf{With proportion of DE gene (pDEG) = 0.02}}                                                                                                                                                                                                                                                                                                                                              \\ \cline{2-8} 
                                  & \multicolumn{1}{l|}{\textbf{TPR}}                       & \multicolumn{1}{l|}{\textbf{TNR}}                       & \multicolumn{1}{l|}{\textbf{FPR}}                       & \multicolumn{1}{l|}{\textbf{FNR}}                       & \multicolumn{1}{l|}{\textbf{MER}}                       & \multicolumn{1}{l|}{\textbf{FDR}}                       & \multicolumn{1}{l|}{\textbf{AUC}}                       \\ \hline
T-test                            & \begin{tabular}[c]{@{}c@{}}0.702\\ (0.932)\end{tabular} & \begin{tabular}[c]{@{}c@{}}0.006\\ (0.001)\end{tabular} & \begin{tabular}[c]{@{}c@{}}0.994\\ (0.999)\end{tabular} & \begin{tabular}[c]{@{}c@{}}0.298\\ (0.068)\end{tabular} & \begin{tabular}[c]{@{}c@{}}0.012\\ (0.003)\end{tabular} & \begin{tabular}[c]{@{}c@{}}0.298\\ (0.068)\end{tabular} & \begin{tabular}[c]{@{}c@{}}0.702\\ (0.932)\end{tabular} \\ \hline
SAM                               & \begin{tabular}[c]{@{}c@{}}0.775\\ (0.935)\end{tabular} & \begin{tabular}[c]{@{}c@{}}0.005\\ (0.001)\end{tabular} & \begin{tabular}[c]{@{}c@{}}0.995\\ (0.999)\end{tabular} & \begin{tabular}[c]{@{}c@{}}0.225\\ (0.065)\end{tabular} & \begin{tabular}[c]{@{}c@{}}0.009\\ (0.003)\end{tabular} & \begin{tabular}[c]{@{}c@{}}0.225\\ (0.065)\end{tabular} & \begin{tabular}[c]{@{}c@{}}0.775\\ (0.935)\end{tabular} \\ \hline
LIMMA                             & \begin{tabular}[c]{@{}c@{}}0.810\\ (0.935)\end{tabular} & \begin{tabular}[c]{@{}c@{}}0.004\\ (0.001)\end{tabular} & \begin{tabular}[c]{@{}c@{}}0.996\\ (0.999)\end{tabular} & \begin{tabular}[c]{@{}c@{}}0.190\\ (0.065)\end{tabular} & \begin{tabular}[c]{@{}c@{}}0.008\\ (0.003)\end{tabular} & \begin{tabular}[c]{@{}c@{}}0.190\\ (0.065)\end{tabular} & \begin{tabular}[c]{@{}c@{}}0.810\\ (0.935)\end{tabular} \\ \hline
Proposed                          & \begin{tabular}[c]{@{}c@{}}0.890\\ (0.935)\end{tabular} & \begin{tabular}[c]{@{}c@{}}0.002\\ (0.001)\end{tabular} & \begin{tabular}[c]{@{}c@{}}0.998\\ (0.999)\end{tabular} & \begin{tabular}[c]{@{}c@{}}0.110\\ (0.050)\end{tabular} & \begin{tabular}[c]{@{}c@{}}0.004\\ (0.002)\end{tabular} & \begin{tabular}[c]{@{}c@{}}0.110\\ (0.050)\end{tabular} & \begin{tabular}[c]{@{}c@{}}0.890\\ (0.950)\end{tabular} \\ \hline
\multirow{2}{*}{\textbf{Methods}} & \multicolumn{7}{c|}{\textbf{With proportion of DE gene (pDEG) = 0.06}}                                                                                                                                                                                                                                                                                                                                              \\ \cline{2-8} 
                                  & \textbf{TPR}                                            & \textbf{TNR}                                            & \textbf{FPR}                                            & \textbf{FNR}                                            & \textbf{MER}                                            & \textbf{FDR}                                            & \textbf{AUC}                                            \\ \hline
T-test                            & \begin{tabular}[c]{@{}c@{}}0.772\\ (0.933)\end{tabular} & \begin{tabular}[c]{@{}c@{}}0.012\\ (0.004)\end{tabular} & \begin{tabular}[c]{@{}c@{}}0.988\\ (0.996)\end{tabular} & \begin{tabular}[c]{@{}c@{}}0.228\\ (0.067)\end{tabular} & \begin{tabular}[c]{@{}c@{}}0.023\\ (0.007)\end{tabular} & \begin{tabular}[c]{@{}c@{}}0.228\\ (0.067)\end{tabular} & \begin{tabular}[c]{@{}c@{}}0.771\\ (0.933)\end{tabular} \\ \hline
SAM                               & \begin{tabular}[c]{@{}c@{}}0.810\\ (0.933)\end{tabular} & \begin{tabular}[c]{@{}c@{}}0.010\\ (0.004)\end{tabular} & \begin{tabular}[c]{@{}c@{}}0.990\\ (0.996)\end{tabular} & \begin{tabular}[c]{@{}c@{}}0.190\\ (0.067)\end{tabular} & \begin{tabular}[c]{@{}c@{}}0.019\\ (0.007)\end{tabular} & \begin{tabular}[c]{@{}c@{}}0.190\\ (0.067)\end{tabular} & \begin{tabular}[c]{@{}c@{}}0.809\\ (0.933)\end{tabular} \\ \hline
IMMA                              & \begin{tabular}[c]{@{}c@{}}0.823\\ (0.933)\end{tabular} & \begin{tabular}[c]{@{}c@{}}0.009\\ (0.004)\end{tabular} & \begin{tabular}[c]{@{}c@{}}0.991\\ (0.996)\end{tabular} & \begin{tabular}[c]{@{}c@{}}0.177\\ (0.067)\end{tabular} & \begin{tabular}[c]{@{}c@{}}0.018\\ (0.007)\end{tabular} & \begin{tabular}[c]{@{}c@{}}0.177\\ (0.067)\end{tabular} & \begin{tabular}[c]{@{}c@{}}0.823\\ (0.933)\end{tabular} \\ \hline
Proposed                          & \begin{tabular}[c]{@{}c@{}}0.911\\ (0.959)\end{tabular} & \begin{tabular}[c]{@{}c@{}}0.005\\ (0.002)\end{tabular} & \begin{tabular}[c]{@{}c@{}}0.995\\ (0.996)\end{tabular} & \begin{tabular}[c]{@{}c@{}}0.089\\ (0.041)\end{tabular} & \begin{tabular}[c]{@{}c@{}}0.009\\ (0.004)\end{tabular} & \begin{tabular}[c]{@{}c@{}}0.089\\ (0.041)\end{tabular} & \begin{tabular}[c]{@{}c@{}}0.911\\ (0.933)\end{tabular} \\ \hline
\end{tabular}
\end{table}

\subsection{Simulated Gene Expression Profiles generated from t- Distribution}
We also investigated the performance of the proposed method in a comparison of other three methods (T-test, SAM and LIMMA) for non-normal case; accordingly we generated 100 simulated datasets from t-distribution with 10 degrees of freedom. We set the mean and variance as before. We estimated different performance measures such as TPR, TNR, FPR, FNR, MER, FDR and AUC based on 20 estimated DE genes by four methods for each of 100 datasets. The average values of performance measures are summarized in Table 2. From this table we notice that the performances of all the methods deteriorated when the datasets came from t-distribution. We also observe that the proposed method performed better than the other three methods (T-test, SAM and LIMMA). For example, the proposed method produces AUC = $0.469$ ($0.887$) which is larger than $0.316$ ($0.830$), $0.326$ ($0.832$) and $0.411$ ($0.880$) for the competitors T-test, SAM and LIMMA. The boxplots in Fig.3 and ROC curve in Fig.1(c-d) also revealed similar results like Table 2. We also noticed from boxplots that the proposed method has less variability among the other three methods. From this analysis we may conclude that the performance of the proposed method has improved than the three well-known gene selection methods.

 \begin{figure}
\begin{center}
\includegraphics[width=8cm, height=8cm]{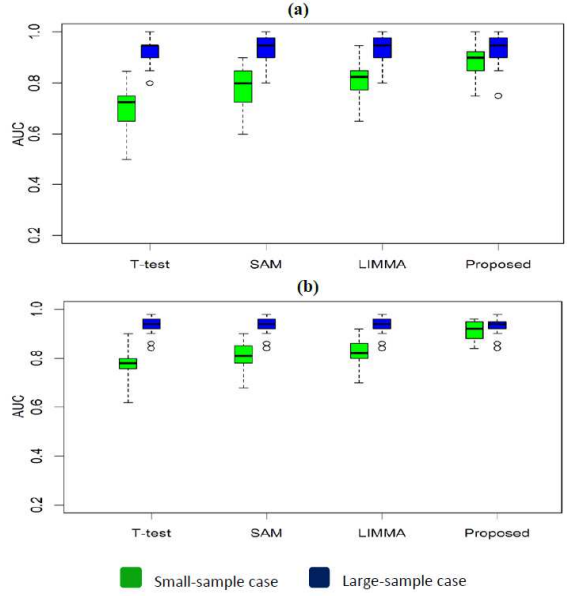}
\caption{ Performance evaluation using boxplot of AUC values produced by the four methods (T-test, SAM, LIMMA and Proposed) based on 100 datasets were taken from normal distribution for small-and large-sample cases (a) Boxplot of AUC values with proportion of DE gene=0.02. (b) Boxplot of AUC values with proportion of DE gene=0.06. Each dataset contains p =1000 genes.}
\label{fig:box1}
\end{center}
\end{figure}

\begin{table}[]
\caption{Performance evaluation of different methods based on simulated gene expression dataset generated from t-distribution}
\begin{tabular}{|l|c|c|c|c|c|c|c|}
\hline
\multicolumn{1}{|c|}{\multirow{2}{*}{\textbf{Methods}}} & \multicolumn{7}{c|}{\textbf{With proportion of DE gene (pDEG) = 0.02}}                                                                                                                                                                                                                                                                                                                                                     \\ \cline{2-8} 
\multicolumn{1}{|c|}{}                                  & \textbf{TPR}                                             & \textbf{TNR}                                             & \textbf{FPR}                                             & \textbf{FNR}                                             & \textbf{MER}                                             & \textbf{FDR}                                             & \textbf{AUC}                                             \\ \hline
T-test                                                  & \begin{tabular}[c]{@{}c@{}}0.318 \\ (0.830)\end{tabular} & \begin{tabular}[c]{@{}c@{}}0.014 \\ (0.003)\end{tabular} & \begin{tabular}[c]{@{}c@{}}0.986\\  (0.997)\end{tabular} & \begin{tabular}[c]{@{}c@{}}0.682 \\ (0.170)\end{tabular} & \begin{tabular}[c]{@{}c@{}}0.027 \\ (0.007)\end{tabular} & \begin{tabular}[c]{@{}c@{}}0.682\\  (0.170)\end{tabular} & \begin{tabular}[c]{@{}c@{}}0.316 \\ (0.830)\end{tabular} \\ \hline
SAM                                                     & \begin{tabular}[c]{@{}c@{}}0.328 \\ (0.832)\end{tabular} & \begin{tabular}[c]{@{}c@{}}0.014\\  (0.003)\end{tabular} & \begin{tabular}[c]{@{}c@{}}0.986 \\ (0.997)\end{tabular} & \begin{tabular}[c]{@{}c@{}}0.672\\  (0.168)\end{tabular} & \begin{tabular}[c]{@{}c@{}}0.027 \\ (0.007)\end{tabular} & \begin{tabular}[c]{@{}c@{}}0.672 \\ (0.168)\end{tabular} & \begin{tabular}[c]{@{}c@{}}0.326 \\ (0.832)\end{tabular} \\ \hline
LIMMA                                                   & \begin{tabular}[c]{@{}c@{}}0.412\\  (0.880)\end{tabular} & \begin{tabular}[c]{@{}c@{}}0.012\\  (0.002)\end{tabular} & \begin{tabular}[c]{@{}c@{}}0.988 \\ (0.998)\end{tabular} & \begin{tabular}[c]{@{}c@{}}0.588 \\ (0.120)\end{tabular} & \begin{tabular}[c]{@{}c@{}}0.024\\ (0.005)\end{tabular}  & \begin{tabular}[c]{@{}c@{}}0.588 \\ (0.120)\end{tabular} & \begin{tabular}[c]{@{}c@{}}0.411\\  (0.880)\end{tabular} \\ \hline
Proposed                                                & \begin{tabular}[c]{@{}c@{}}0.470 \\ (0.888)\end{tabular} & \begin{tabular}[c]{@{}c@{}}0.011 \\ (0.002)\end{tabular} & \begin{tabular}[c]{@{}c@{}}0.988 \\ (0.998)\end{tabular} & \begin{tabular}[c]{@{}c@{}}0.530\\ (0.112)\end{tabular}  & \begin{tabular}[c]{@{}c@{}}0.021\\  (0.004)\end{tabular} & \begin{tabular}[c]{@{}c@{}}0.530 \\ (0.112)\end{tabular} & \begin{tabular}[c]{@{}c@{}}0.469 \\ (0.887)\end{tabular} \\ \hline
\end{tabular}
\end{table}

 \begin{figure}
\begin{center}
\includegraphics[width=8cm, height=8cm]{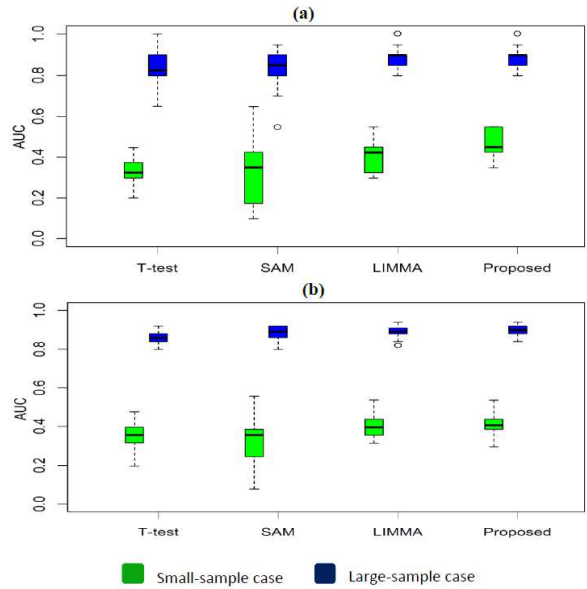}
\caption{ Performance evaluation using boxplot of AUC values produced by the four methods (T-test, SAM, LIMMA and Proposed) based on 100 datasets were taken from t-distribution distribution for small-and large-sample cases (a) Boxplot of AUC values with proportion of DE gene=0.02. (b) Boxplot of AUC values with proportion of DE gene=0.06. Each dataset contains p =1000 genes.}
\label{fig:box2}
\end{center}
\end{figure}

\subsection{Application to colon cancer microarray data}
The  data consist of expression levels of 2000
genes obtained from a microarray study on 62 colon tissue  samples collected from colon-cancer patients. Among $62$ colon tissue, $40$ tumor tissues, coded 2 and 22 normal tissues, coded 1 \cite{Alon-99}.
The goal here is to characterize the underlying interactions between genetic makers for their association with the  colon-cancer patients and the healthy persons.

\begin{figure}
\begin{center}
\includegraphics[width=8cm, height=8cm]{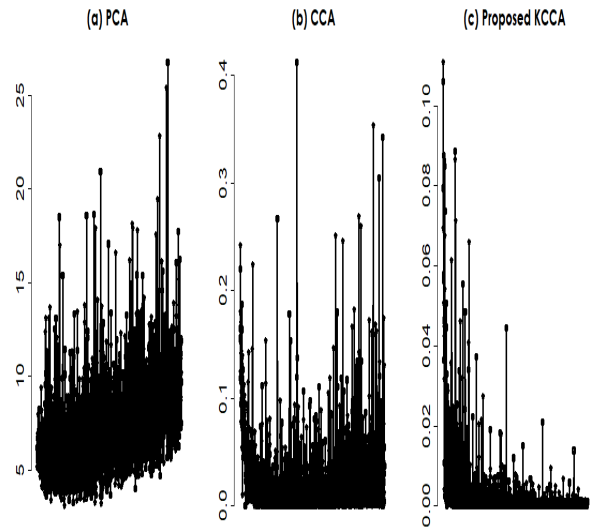}
\caption{ The influence value of genes using three methods: principal components analysis (PCOut), linear canonical correlation
analysis (LCCA), and kernel canonical correlation analysis (KCCA).}
\label{fig:ipon}
\end{center}
\end{figure}

\begin{figure}
\begin{center}
\includegraphics[width=8cm, height=8cm]{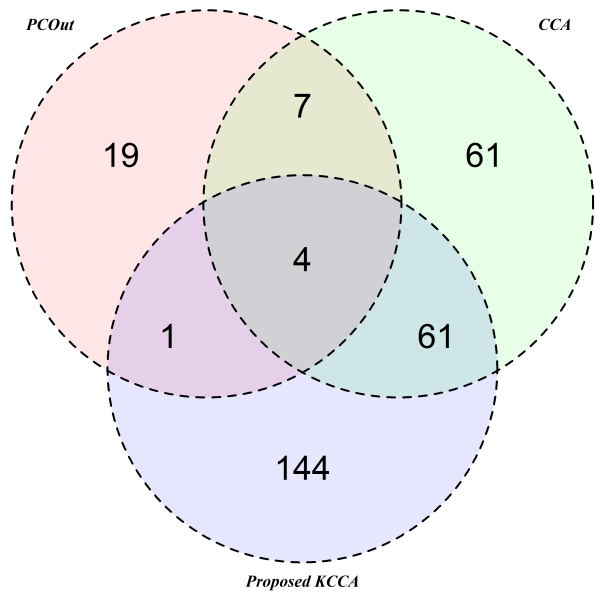}
\caption{ The Venn diagram of the selected genes using three methods: principal components analysis (PCOut), linear canonical correlation
analysis (LCCA), and kernel canonical correlation analysis (KCCA).}
\label{fig:venn}
\end{center}
\end{figure}

 To calculate the influence value of  each gene, we used three methods: PCAout, liner CCA and the proposed method, KCCA, respectively.  Figure \ref{fig:ipon} visualizes the plots of  absolute influence value  for  $2000$  genes.
 By  the outliers detection technique  in the  one dimensional  influence value of each method,  we obtained $31$, $133$ and $210$ genes using PCAout, liner CCA and the proposed method KCCA, respectively. To compare the selected genes, we made a Venn-diagram of the selected genes from the three methods. Figure \ref{fig:venn}  presents the Venn-diagram of the PCOut, LCCAOut, and KCCAOut methods. From this figure, we observe that the disjoint selected genes of PCOut, LCCAOut, and KCCAOut are $19$, $61$, and $144$, respectively. The number of common genes between PCOut and LCCAOut, and PCOut and KCCAOut, and LCCAOut and KCCAOut are 7, 1, and 61, respectively. All methods selected 4 common genes: J00231, T57780, M94132 and M87789.

\begin{figure}
\begin{center}
\includegraphics[width=9cm, height=12cm]{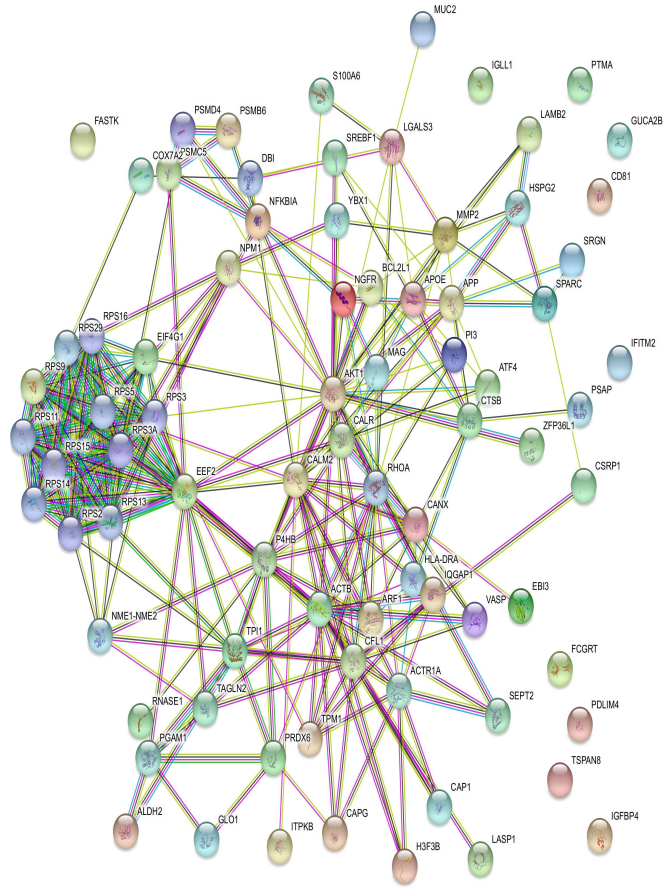}
\caption{ The network of the selected genes.}
\label{fig:ggn1}
\end{center}
\end{figure}

Genes do not function alone; rather, they interact  with each other. When genes share a similar set of GO annotation terms, they are most likely to be involved with similar biological mechanisms. To verify this,  we extracted the gene-gene  networks using STRING \cite{STRING-15}. STRING imports protein association knowledge from databases of both physical interactions and curated biological pathways. In STRING, the simple interaction unit is the functional relationship between two proteins/genes that can contribute to a common biological purpose. Figure~ \ref{fig:ggn1}  shows the gene-gene network based on the  protein interactions between the  combined $210$.  In this figure, the color saturation of the edges represents the confidence score of a functional association. Further network analysis shows that the number of nodes, number of edges, average node degree, clustering coefficient, PPI enrichment $p$-values are $75$, $214$, $5.71$, $0.473$ for $p\leq 8.22\times 10^{-15}$, respectively. This network of genes has significantly more interactions than expected, which indicates that they may function in a concerted effort.

\section{Concluding remarks}
\label{Sec:con}
The kernel based methods provide more powerful and reproducible outputs, but the interpretation of the results remains challenging. Incorporating biological knowledge information (e.g., GO) can provide additional evidences on the results. The performance of the proposed method was evaluated  on both simulated  and  a real  data. The extensive simulation studies show the power gain of the proposed  method relative to the alternative methods.

The utility of the proposed method is further demonstrated with the application to colon cancer microarray data. According to the  influence values, the proposed method is able to  rank   the  influence of a gene and the genes are  are identified to be highly related to  disease. Using a ourlier detection methods the proposed method extracts the $210$ genes out of $2000$ genes, which are considered to have significant impact on   the patients. By  conducting gene ontology, pathway analysis, and network analysis including visualization, we find evidences that the selected genes have a significant influence on the manifestation of colon cancer disease and can serve as a distinct feature for the classification of colon cancer patients from the healthy controls.

 Although  the Gaussian  kernel has a free parameter (bandwidth), in this study,  we used the median of the  pairwise distance as the bandwidth for the  Gaussian kernel, which appears to be practical.  In future work, tt must be emphasized that choosing a suitable kernel is indispensable. 
\subsection*{Acknowledgments}
 The authors wish to thank the University Grants Commission of Bangladesh for         support.                                                                                
\begingroup
\bibliographystyle{apacite}
\bibliography{Ref-UKIF}
\endgroup                                                                                                         
\end{document}